\documentclass{article}

\usepackage[preprint]{neurips_2026}

\usepackage{graphicx}
\usepackage{amsmath}
\usepackage{amsthm}
\usepackage{subcaption}
\usepackage{algorithm}
\usepackage{algorithmic}

\usepackage[utf8]{inputenc} 
\usepackage[T1]{fontenc}    
\usepackage{hyperref}       
\usepackage{url}            
\usepackage{booktabs}       
\usepackage{amsfonts}       
\usepackage{nicefrac}       
\usepackage{microtype}      
\usepackage{xcolor}         

\title{When Backdoors Meet Partial Observability: Attacking Real-World Reinforcement Learning}

\author{%
 Tairan Huang
  \quad Qingqing Ye
  \quad Yulin Jin
  \quad Jiawei Lian
  \\ \textbf{Yaxin Xiao
  \quad Yi Wang
  \quad Haibo Hu }
  \\
  Department of Electrical and Electronic Engineering \\
  Hong Kong Polytechnic University \\
  Hong Kong SAR, China \\
  \texttt{tairan.huang@connect.polyu.hk}
  \quad \texttt{qqing.ye@polyu.edu.hk}
  \\ \texttt{\{yulin.jin, jiawei.lian, 20034165r\}@connect.polyu.hk}
  \\ \texttt{\{yi-eie.wang, haibo.hu\}@polyu.edu.hk}
}

\begin{document}

\maketitle

\begin{abstract}
Backdoor attacks can cause reinforcement learning (RL) policies to behave normally under clean inputs while executing malicious behaviors when triggers are present. 
Existing RL backdoor attacks are primarily studied in simulation and often assume that attackers can reliably manipulate the observations driving policy decisions. 
This assumption becomes fragile in real-world deployment, where RL policies commonly rely on multimodal observations. 
Attackers can manipulate visual inputs through physical triggers, but auxiliary states such as LiDAR and odometry signals remain uncontrollable and vary across trajectories. 
We study this overlooked challenge and propose a diffusion-guided backdoor attack framework (DGBA) for real-world RL. 
DGBA uses small printable visual patches as triggers and learns a stochastic trigger distribution via conditional diffusion to maintain consistent attack activation under varying uncontrollable states. 
We further introduce an advantage-based poisoning strategy that injects triggers only at decision-critical training states. 
Experiments on a physical TurtleBot3 platform show that DGBA consistently outperforms prior RL backdoor attacks while preserving normal task performance. 
\textbf{Demo videos and code are available
in the supplementary material.}
\end{abstract}

\section{Introduction}

Reinforcement learning (RL) has moved beyond simulation and is increasingly deployed on physical robots for navigation, manipulation, and autonomous control~\cite{DBLP:conf/aaai/TangAHCM025,DBLP:conf/aaai/WangCLWP24,DBLP:conf/iclr/ZhuY0SHSKL20,DBLP:conf/iros/ChandraKMSB25}. 
In real-world deployment, RL agents often rely on multimodal observations that combine visual inputs with auxiliary sensor signals. 
This shift introduces new security challenges that are largely absent in standard simulation settings.

Backdoor attacks pose a serious threat to RL systems. 
A backdoored policy behaves normally under clean inputs but executes malicious behaviors when triggers are present. 
Recent studies show that such attacks can achieve high success rates in simulation even under small poisoning budgets~\cite{trojdrl,badrl,sleepernets}. 
However, existing RL backdoor attacks typically assume that attackers can reliably manipulate the observations driving policy decisions. 
This assumption becomes fragile in real-world deployment.

In deployed robotic systems, the policy observation often consists of visual observations and auxiliary sensor states. 
For example, a mobile robot may jointly use camera images and LiDAR signals for decision making. 
An attacker can physically manipulate the visual channel by placing triggers in the environment, but auxiliary sensor states remain uncontrollable and vary across trajectories. 
As a result, the same visual trigger may correspond to different overall observations, making deterministic backdoor triggers significantly less reliable in real-world deployment.

This raises a central question: when attackers can control only part of the observation space, how can backdoor attacks remain effective? 
A natural solution is to move from deterministic triggers to stochastic trigger distributions. 
Instead of learning a single fixed trigger pattern, the attacker should learn a family of triggers that consistently activate malicious behaviors under varying uncontrollable states.

To address this challenge, we propose a diffusion-guided backdoor attack framework (DGBA) for real-world RL. 
DGBA uses small printable floor patches as physical triggers. 
A conditional diffusion model learns a stochastic distribution over trigger appearances, enabling consistent activation under varying environmental and sensor conditions~\cite{ddpm,ddim,cddm}. 
We further introduce an advantage-based poisoning strategy that injects triggers only at decision-critical training states to improve poisoning efficiency.

We evaluate DGBA on a TurtleBot3 mobile robot~\cite{tb3}, a widely used real-world platform for reinforcement learning research~\cite{logo,rnac,fedora,emrld}. 
Experiments show that DGBA consistently outperforms prior RL backdoor attacks while preserving normal task performance in real-world deployment.

\paragraph{Contributions.}
Our main contributions are:

\begin{itemize}
    \item We identify a previously overlooked challenge in real-world RL backdoor attacks: attackers can manipulate only part of multimodal observations while auxiliary states remain uncontrollable.

    \item We propose DGBA, a diffusion-guided backdoor framework that learns stochastic printable triggers for real-world RL systems.

    \item We introduce an advantage-based poisoning strategy that injects triggers at decision-critical states under limited poisoning budgets.

    \item We demonstrate on a real TurtleBot3 platform that DGBA consistently outperforms existing RL backdoor attacks.
\end{itemize}
\section{Related Work}

\subsection{Backdoor Attacks in Reinforcement Learning}

Backdoor attacks in reinforcement learning (RL) are studied as a training-time poisoning threat, 
where a policy learns hidden malicious behaviors that are activated by triggers at test time, while behaving normally otherwise.
TrojDRL demonstrates that inserting triggered samples and modifying reward signals during training can induce targeted misbehavior in deep RL policies~\cite{trojdrl}. 

Subsequent work focuses on improving attack efficiency and reducing poisoning budgets. 
BadRL observes that only a subset of training states strongly influences policy updates, and proposes selecting the top-ranked critical states for sparse poisoning, 
achieving high attack success with significantly fewer poisoned samples~\cite{badrl}. 
SleeperNets formulates poisoning as an outer-loop attack, where the adversary observes completed training trajectories and then selects critical states and rewards to poison before policy updates, enabling data-efficient backdoor attacks across RL algorithms~\cite{sleepernets}.

Despite these advances, existing methods are primarily developed and evaluated in simulated environments. 
Most experiments rely on Atari-style benchmarks~\cite{atari} or simplified control settings where observations are fully controllable by the attacker. 
They largely overlook real-world RL systems that rely on multimodal observations, where attackers can manipulate only part of the observation space while auxiliary sensor states remain uncontrollable.

\subsection{Real-World Reinforcement Learning}

Recent progress has demonstrated that reinforcement learning policies can be successfully deployed on physical mobile robots. 
TurtleBot3 has emerged as a widely adopted real-world platform for evaluating RL algorithms due to its accessibility and standardized ROS-based control interface.

LOGO introduces a policy optimization framework that leverages sub-optimal behavior policies to guide RL training under sparse rewards, and validates waypoint tracking and obstacle avoidance policies on a TurtleBot3 in real-world experiments~\cite{logo}. 
RNAC proposes a natural actor-critic method to learn policies resilient to uncertain dynamics, and validates real-world deployment on a TurtleBot3 robot in a navigation task~\cite{rnac}.
FEDORA studies federated offline reinforcement learning, using ensemble aggregation to combine knowledge from distributed clients. 
It evaluates real-world waypoint navigation with obstacle avoidance on a TurtleBot3 robot~\cite{fedora}.
EMRLD studies meta-RL in sparse-reward settings by leveraging demonstration data from sub-optimal agents to guide policy adaptation. 
It validates real-world navigation experiments on a TurtleBot3 robot under environmental drift~\cite{emrld}.

Collectively, these works establish TurtleBot3 as a representative and practical testbed for real-world reinforcement learning. 
Meanwhile, backdoor attacks in RL have been studied almost exclusively in simulation. 
The intersection between real-world RL deployment and backdoor security vulnerabilities, especially under multimodal observations with partially controllable inputs, remains largely unexplored.

\begin{figure}[t]
  \centering
  \includegraphics[width=0.98\textwidth]{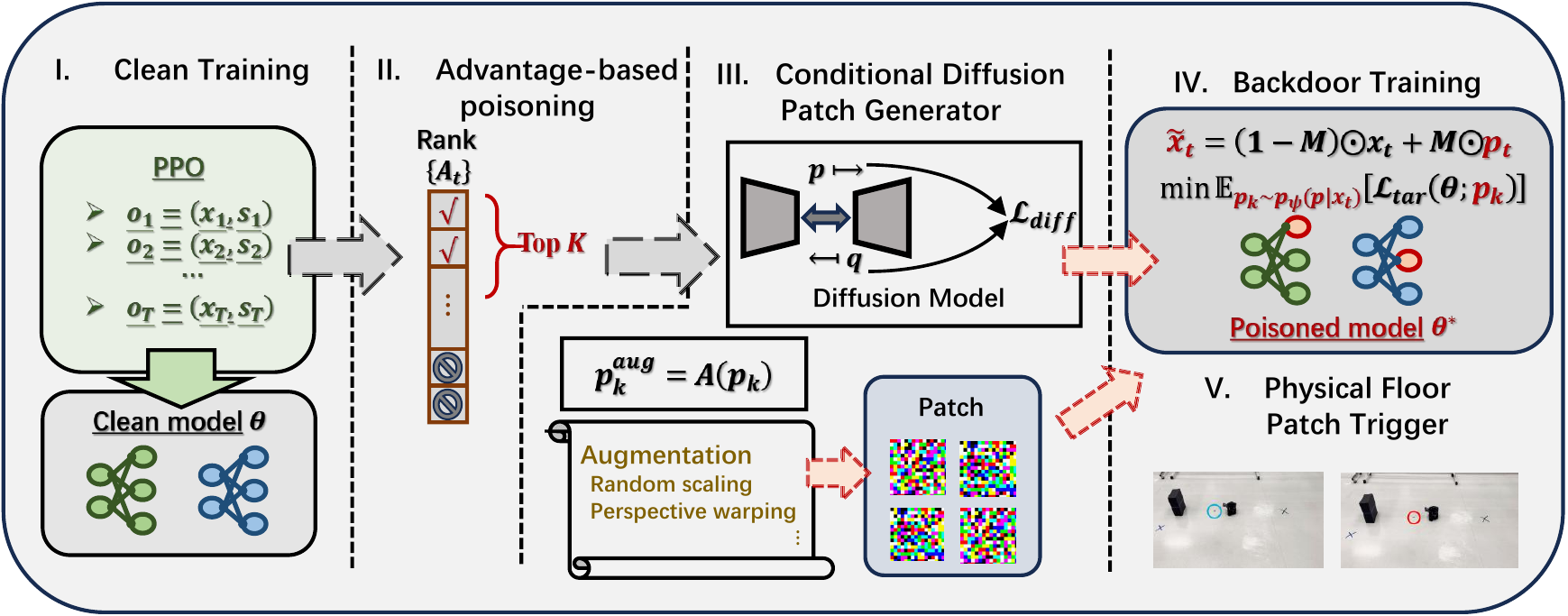}
  \caption{Overview of DGBA backdoor training and deployment pipeline.}
  \label{fig:dgba_overview}
\end{figure}

\section{Problem Setting}

\subsection{Victim Model}

We consider a reinforcement learning policy $\pi_\theta$ deployed on a physical mobile robot. 
At each time step $t$, the robot receives a multimodal observation
\begin{equation}
o_t=(x_t,s_t),
\end{equation}
where $x_t$ denotes visual observations from an onboard camera and $s_t$ represents auxiliary sensor states such as LiDAR signals. 
The policy outputs a continuous control action
\begin{equation}
a_t=\pi_\theta(o_t).
\end{equation}

Unlike standard simulated benchmarks, real-world RL policies often rely on multiple sensing modalities for decision making. 
This creates additional constraints for attackers, since not all components of the observation are equally controllable.

\subsection{Threat Model and Attack Objective}

We study training-time backdoor attacks under a constrained poisoning setting. 
The attacker cannot modify the policy architecture, reward function, or auxiliary sensors. 
Instead, the attacker can only manipulate a small fraction of training data by injecting visual triggers.

A trigger function $\tau(\cdot)$ modifies only the visual observation:
\begin{equation}
\tilde{o}_t=(\tilde{x}_t,s_t),
\end{equation}
where
\begin{equation}
\tilde{x}_t=\tau(x_t).
\end{equation}

The auxiliary state $s_t$ remains unchanged and uncontrollable.

The attack objective is to learn a policy that behaves normally on clean observations,
\begin{equation}
\pi_\theta(o_t)\approx \pi_\theta^{\text{clean}}(o_t),
\end{equation}
while producing attacker-specified behavior when the trigger is present,
\begin{equation}
\pi_\theta(\tilde{o}_t)\rightarrow a_t^{\text{target}}.
\end{equation}

The attacker aims to achieve high attack success under limited poisoning budgets while preserving clean-task performance.

\subsection{Challenge of Partial Observation Control}

A key challenge in real-world deployment is that attackers can manipulate only part of the observation space. 
While visual inputs can be modified through physical triggers, auxiliary sensor states vary across trajectories and remain outside attacker control.

As a result, the same visual trigger may correspond to different multimodal observations:
\begin{equation}
(\tilde{x}_t,s_t^{(1)}) \neq (\tilde{x}_t,s_t^{(2)}).
\end{equation}

This makes deterministic triggers significantly less reliable in real-world RL systems. 
Our goal is to design backdoor attacks that remain effective under this partial-control setting without requiring access to auxiliary sensing modalities.

\section{Method}
\label{sec:method}

\subsection{Overview}
\label{sec:method_overview}

We propose DGBA, a diffusion-guided backdoor attack framework for real-world reinforcement learning under partial observation control. 
DGBA consists of a physical patch trigger (Sec.~\ref{sec:patch}), a conditional diffusion generator (Sec.~\ref{sec:diffusion}), physical-style augmentation (Sec.~\ref{sec:aug}), advantage-based poisoning (Sec.~\ref{sec:advpoison}), and a unified training and deployment pipeline (Sec.~\ref{sec:pipeline}). 
Figure~\ref{fig:dgba_overview} summarizes the overall framework.

\subsection{Physical Floor Patch Trigger}
\label{sec:patch}

We implement the backdoor trigger as a small printable patch placed on the floor in the robot workspace. 
At time step $t$, the observation is $o_t=(x_t,s_t)$, where $x_t\in\mathbb{R}^{H\times W\times C}$ is an RGB image and $s_t$ denotes auxiliary sensor states. 

A binary mask $M\in\{0,1\}^{H\times W}$ specifies the trigger region. 
Given a patch pattern $p_t$, the triggered visual observation is
\begin{equation}
\tilde{x}_t=(1-M)\odot x_t+M\odot p_t.
\end{equation}

The triggered policy input becomes
\begin{equation}
\tilde{o}_t=(\tilde{x}_t,s_t).
\end{equation}

The trigger modifies only a small visual region while preserving the rest of the scene. 
The patch is physically printable and can be directly deployed in real environments.

\subsection{Conditional Diffusion Patch Generator}
\label{sec:diffusion}

A deterministic trigger is often unreliable in real-world deployment. 
Although attackers can manipulate visual observations through physical patches, auxiliary sensor states remain uncontrollable and vary across trajectories. 
As a result, the same visual trigger may correspond to different multimodal observations, making fixed trigger patterns unstable.

To address this challenge, we model the trigger as a stochastic distribution rather than a single deterministic pattern. 
We adopt a denoising diffusion probabilistic model to generate trigger samples.

Let $p_0$ denote a clean patch image. 
The forward diffusion process corrupts $p_0$ with Gaussian noise:
\begin{equation}
q(p_k|p_0)=
\mathcal{N}
\left(
p_k;
\sqrt{\bar{\alpha}_k}p_0,
(1-\bar{\alpha}_k)I
\right).
\end{equation}

The cumulative noise schedule is defined as
\begin{equation}
\bar{\alpha}_k=\prod_{i=1}^{k}\alpha_i.
\end{equation}

The reverse denoising process is parameterized by a neural network $\epsilon_\psi$, which predicts injected noise conditioned on both the corrupted patch and visual observation $x_t$. 
The training objective is
\begin{equation}
\mathcal{L}_{\text{diff}}(\psi)=
\mathbb{E}_{k,p_0,\epsilon}
\left[
\|
\epsilon-\epsilon_\psi(p_k,x_t,k)
\|^2
\right].
\end{equation}

After training, trigger samples are generated through
\begin{equation}
p_k\sim p_\psi(p|x_t).
\end{equation}

During poisoning, sampled triggers are injected into training trajectories. 
This trains the policy to associate a distribution of trigger appearances with malicious behaviors instead of relying on a single fixed trigger. 

Because auxiliary states remain uncontrollable, learning a stochastic trigger distribution improves attack consistency under varying multimodal observations.

\subsection{Physical-Style Patch Augmentation}
\label{sec:aug}

Diffusion sampling provides trigger diversity, but sim-to-real discrepancies still exist due to camera geometry, lighting conditions, and environmental clutter. 

To bridge this gap, we apply physical-style augmentation:
\begin{equation}
p_t^{aug}=\mathcal{A}(p_t).
\end{equation}

The augmentation operator includes random scaling, perspective warping, in-plane rotation, brightness and contrast jitter, color perturbation, and partial occlusion.

These transformations simulate realistic visual variations encountered during deployment and improve trigger consistency in physical environments.

\subsection{Advantage-Based Poisoning}
\label{sec:advpoison}

The attacker can poison only a small fraction of training data. 
We therefore inject triggers only at decision-critical states that have the largest influence on policy updates.

For each transition, PPO computes an advantage estimate
\begin{equation}
A_t=\hat{R}_t-V_\phi(o_t).
\end{equation}

Transitions with larger $|A_t|$ contribute more strongly to policy learning. 
We rank rollout steps by $|A_t|$ and select the top
\begin{equation}
K=\lfloor \beta H \rfloor
\end{equation}
steps, where $\beta\ll1$ is the poisoning ratio and $H$ is rollout length.

Only these selected transitions are poisoned, while all remaining samples stay clean.

\paragraph{Target objective.}

Under the trigger, the attacker aims to induce a predefined malicious behavior. 
We optimize
\begin{equation}
\mathcal{L}_{\text{tar}}(\theta)=
\mathbb{E}_{t\in\mathcal{I}}
\left[
\ell(
\pi_\theta(\tilde{o}_t),
a_t^{\text{target}}
)
\right].
\end{equation}

During training, we maximize $\mathcal{L}_{\text{tar}}$ on poisoned samples while optimizing the standard PPO objective on clean samples.

In our experiments, $a_t^{\text{target}}$ corresponds to right-turn, left-turn, straight-driving toward obstacles, or full-stop behaviors.

\subsection{Training and Deployment Pipeline}
\label{sec:pipeline}

DGBA follows a three-stage pipeline.

First, we train a clean PPO policy to obtain strong nominal navigation performance.

Second, we finetune the policy using mixed clean and poisoned trajectories. 
At each iteration, decision-critical states are selected, diffusion-generated triggers are sampled, physical augmentations are applied, and poisoned updates are performed.

Finally, the learned trigger is printed and deployed in the real world. 
The robot observes the trigger through its onboard camera and executes the corresponding malicious behavior.

We evaluate both attack success and clean-task performance in real-world deployment.

\section{Why Does DGBA Work?}
\label{sec:why}

We provide a simple theoretical intuition for why stochastic triggers are more effective than deterministic triggers under partial observation control.

Recall that the policy observes
\begin{equation}
o_t=(x_t,s_t),
\end{equation}
where the attacker can modify only $x_t$ while $s_t$ remains uncontrollable.

For a deterministic trigger $\tau(\cdot)$, the poisoned objective can be written as
\begin{equation}
\mathcal{L}_{\text{det}}
=
\mathbb{E}_{(x_t,s_t)}
\left[
\ell\big(
\pi_\theta(\tau(x_t),s_t),
a_t^{\text{target}}
\big)
\right].
\end{equation}

Because $s_t$ varies across trajectories, the same trigger may be paired with significantly different auxiliary states. 
This creates high variance in the poisoned objective and reduces trigger consistency during deployment.

DGBA instead learns a stochastic trigger distribution $p_\psi(p|x_t)$ and optimizes
\begin{equation}
\mathcal{L}_{\text{sto}}
=
\mathbb{E}_{(x_t,s_t)}
\mathbb{E}_{p\sim p_\psi(p|x_t)}
\left[
\ell\big(
\pi_\theta(\tau_p(x_t),s_t),
a_t^{\text{target}}
\big)
\right]
.
\end{equation}

By optimizing over a trigger distribution rather than a fixed trigger, DGBA minimizes expected attack loss across varying auxiliary states instead of overfitting to specific training trajectories.

As a result, stochastic optimization produces solutions that are less sensitive to shifts in uncontrollable states:
\begin{equation}
\mathrm{Var}_{s_t}
\big[
\mathcal{L}_{\text{sto}}
\big]
\leq
\mathrm{Var}_{s_t}
\big[
\mathcal{L}_{\text{det}}
\big].
\end{equation}

This explains why deterministic triggers often fail when deployed under changing multimodal observations, while DGBA maintains more consistent activation in real-world environments.
\section{Experimental Setup}

\subsection{TurtleBot3 Platform and Multimodal Observations}

We evaluate DGBA on a TurtleBot3 Burger robot equipped with a forward-facing RGB camera and a 2D LiDAR sensor.
At each timestep, the policy receives multimodal observations consisting of visual input $x_t$ and auxiliary sensor input $s_t$.
The attacker can modify only the visual channel through physical triggers, while the LiDAR observations remain untouched.

The robot uses an external USB RGB camera capturing $640\times480$ images at 22--24 Hz, resized to $84\times84$ as visual input $x_t$.
The 2D LiDAR provides full 360$^\circ$ scans at approximately 7 Hz.
Raw scans contain 360 range measurements, which are uniformly subsampled into a 36-dimensional vector $s_t$ for policy input.
All experiments are conducted in an indoor environment with static obstacles.

\subsection{Navigation Task and Evaluation Protocol}

The navigation task requires the robot to reach a predefined goal while avoiding obstacles.
Clean-task performance is measured by the success rate within a fixed time limit.

For attack evaluation, a trigger instance $p^\star$ is sampled from the diffusion generator, exported as a printable RGB image, 
and placed along the navigation route.
Without the trigger patch, the robot follows the clean policy and reaches the goal safely.
When the trigger is present, the backdoored policy executes attacker-specified behaviors (left, right, straight, or stop).
Experiments are repeated under randomized initial orientations to ensure reliable evaluation.

\subsection{Implementation Details}

We use PPO~\cite{ppo} as the primary victim algorithm and additionally evaluate TRPO~\cite{trpo} to test cross-algorithm generalization.
The policy network consists of a convolutional visual encoder processing $3\times84\times84$ images, followed by a two-layer MLP fusing visual features with the 36-dimensional LiDAR input to produce continuous control commands.
The value network shares the same visual encoder.

The diffusion-based trigger generator is a conditional U-Net~\cite{unet} producing $16\times16$ trigger patches at the policy input resolution.
For deployment, patches are upsampled and exported as $128\times128$ printable RGB images.
All policy and diffusion training are conducted in Gazebo~\cite{gazebo} with camera and LiDAR settings matching the real robot.
Trained policies are then evaluated on a physical TurtleBot3 Burger platform.

The diffusion model is trained jointly with the policy using the standard noise-prediction objective for 200 finetuning iterations.
The advantage-based poisoning ratio is set to $\beta=0.05$, modifying only the top 5\% of decision-critical transitions.
The patch mask covers a $16\times16$ region in the lower part of the resized camera image, corresponding to a floor patch region in the robot’s field of view.

Physical-style augmentation includes random scaling, perspective warping, brightness and contrast jitter, color perturbation, and partial occlusion.
These augmentations are applied during finetuning and introduce moderate geometric and photometric variability.
TrojDRL, BadRL, and SleeperNets are trained with the same physical-style data augmentation as DGBA.

Throughout all experiments, attackers modify only visual observations while auxiliary LiDAR inputs remain clean, following the partial observation threat model.

\begin{table}[t]
\centering

\begin{minipage}[t]{0.48\linewidth}
\centering
\caption{Main results on PPO victim under real-world deployment.}
\label{tab:ppo_main}
\begin{tabular}{lcc}
\toprule
Method & CSR (\%) & ASR (\%) \\
\midrule
Clean PPO & 91.1 & -- \\
TrojDRL~\cite{trojdrl} & 85.6 & 34.5 \\
BadRL~\cite{badrl} & 87.3 & 57.0 \\
SleeperNets~\cite{sleepernets} & 88.7 & 21.3 \\
DGBA (ours) & \textbf{89.1} & \textbf{83.5} \\
\bottomrule
\end{tabular}
\end{minipage}
\hfill
\begin{minipage}[t]{0.48\linewidth}
\centering
\caption{Results on TRPO victim under real-world deployment.}
\label{tab:trpo_main}
\begin{tabular}{lcc}
\toprule
Method & CSR (\%) & ASR (\%) \\
\midrule
Clean TRPO & 93.4 & -- \\
TrojDRL~\cite{trojdrl} & 82.5 & 43.9 \\
BadRL~\cite{badrl} & 88.2 & 46.5 \\
SleeperNets~\cite{sleepernets} & 87.9 & 11.9 \\
DGBA (ours) & \textbf{90.7} & \textbf{76.3} \\
\bottomrule
\end{tabular}
\end{minipage}

\end{table}

\section{Results}

\subsection{Main Results on PPO Victim}

We first evaluate DGBA and all baselines on the PPO victim policy under real-world deployment.
Each method is trained with the same poisoning budget ($\beta=5\%$), 
identical network architecture, data augmentation, and deployment setting.
During real-world evaluation, the trained policy is deployed on the TurtleBot3 platform with multimodal observations from camera and LiDAR sensors.

We report two metrics:
(i) \emph{Clean Success Rate (CSR)}, 
defined as the fraction of trials where the robot reaches the goal without collision in the absence of the trigger; 
(ii) \emph{Attack Success Rate (ASR)}, defined as the fraction of trials where the robot executes the target behavior after observing the trigger patch.

For completeness, PPO optimizes 
the clipped surrogate objective
\begin{equation}
\mathcal{L}_{\text{PPO}}(\theta)
=
\mathbb{E}_t
\Big[
\min\!\big(
r_t(\theta) A_t,\,
\bar{r}_t(\theta) A_t
\big)
\Big],
\end{equation}
where
\begin{equation}
r_t(\theta)=\frac{\pi_\theta(a_t\mid o_t)}{\pi_{\theta_{\text{old}}}(a_t\mid o_t)},
\end{equation}
\begin{equation}
\bar{r}_t(\theta)=\text{clip}\!\big(r_t(\theta),1-\epsilon,1+\epsilon\big),
\end{equation}
and $A_t$ is the advantage estimate.

Table~\ref{tab:ppo_main} summarizes the real-world performance.
The clean PPO policy achieves a high success rate, indicating stable navigation in the physical environment.
All backdoored policies preserve comparable clean-task performance, showing that sparse poisoning does not degrade nominal behavior.

In contrast, existing RL backdoor baselines show limited effectiveness when deployed on real robots with multimodal observations.
SleeperNets~\cite{sleepernets} yields the lowest ASR (21.3\%), indicating that representation-level backdoors are sensitive to real-world observation variations.
TrojDRL~\cite{trojdrl} achieves moderate attack performance (34.5\%), but its fixed trigger pattern remains vulnerable to physical imaging variations.
BadRL~\cite{badrl} improves attack effectiveness by targeting decision-critical states for poisoning and reaches an ASR of 57.0\%, 
yet its deterministic trigger remains susceptible to uncontrollable auxiliary sensor states.

DGBA consistently outperforms all baselines in attack success.
Its high ASR of 83.5\% indicates that diffusion-generated physical patch distributions produce stable trigger effects under real-world multimodal observations.
At the same time, DGBA maintains a competitive CSR of 89.1\% among all attack methods, showing that reliable backdoor activation is achieved without sacrificing nominal navigation performance.

\begin{figure}[t]
\centering

\includegraphics[width=0.19\textwidth]{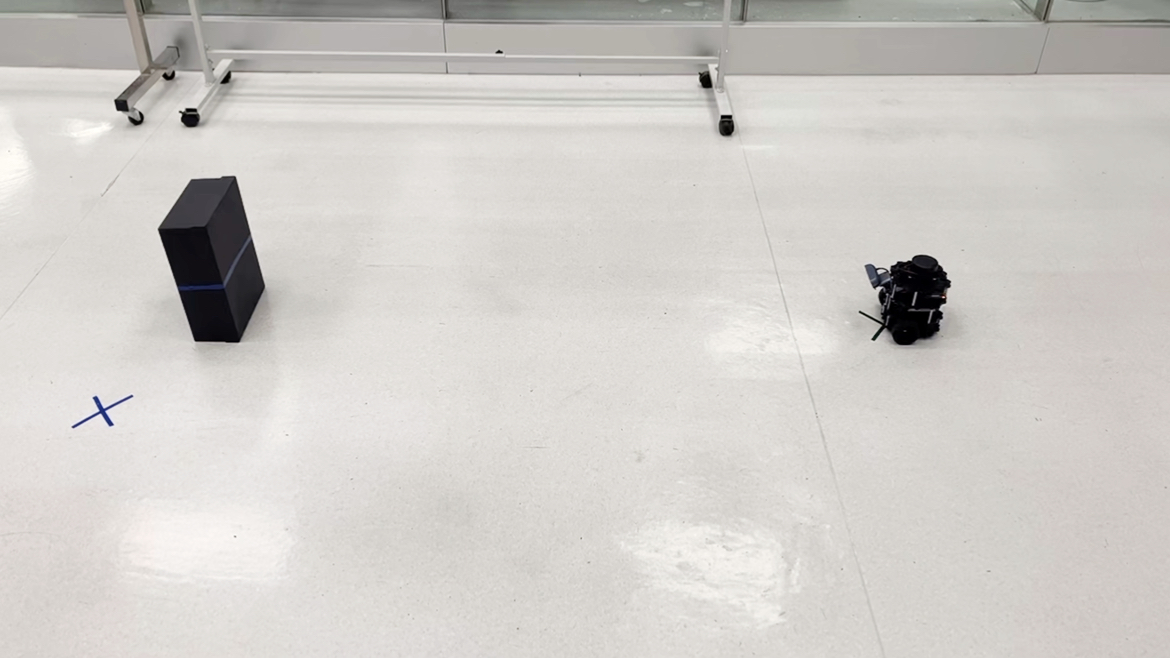}
\includegraphics[width=0.19\textwidth]{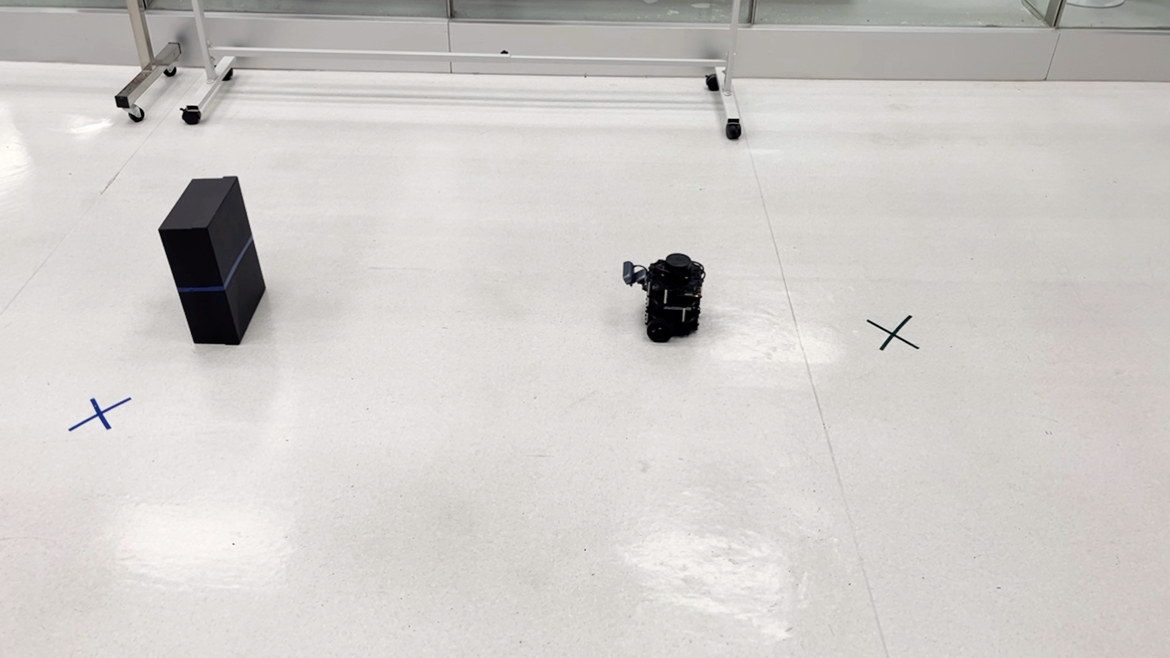}
\includegraphics[width=0.19\textwidth]{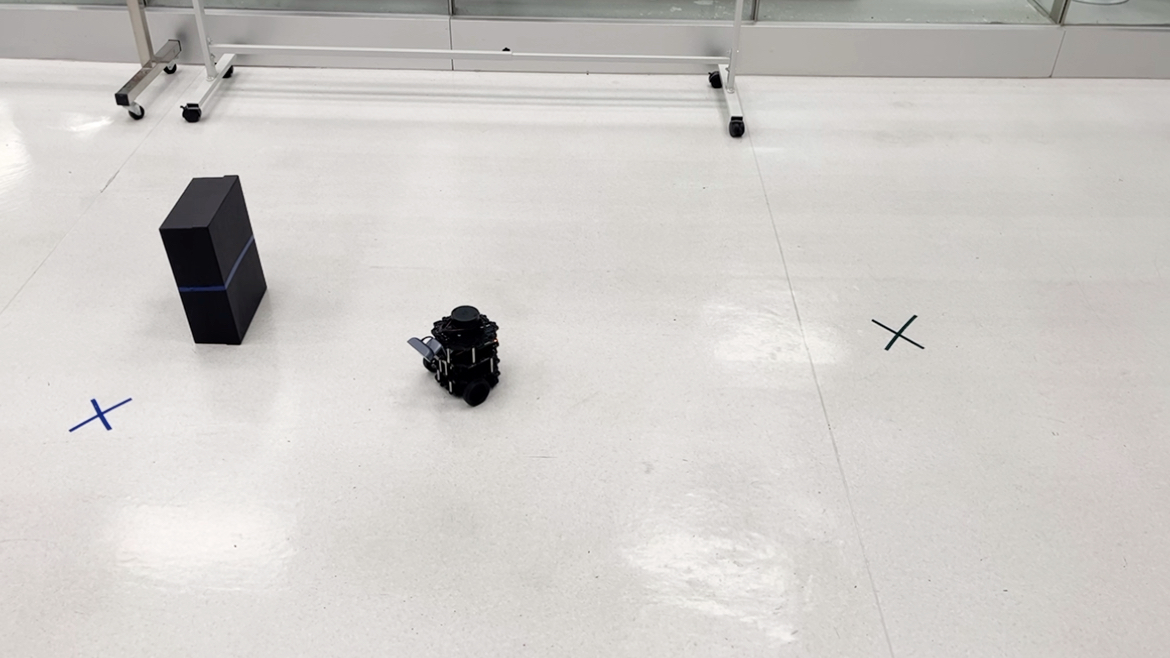}
\includegraphics[width=0.19\textwidth]{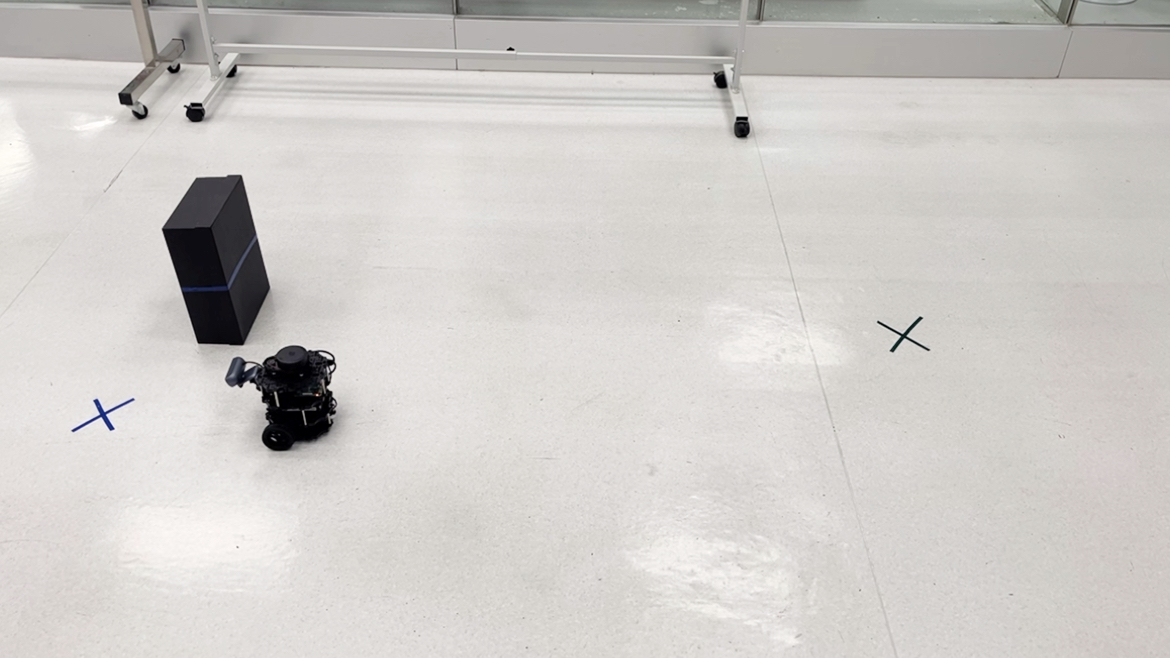}
\includegraphics[width=0.19\textwidth]{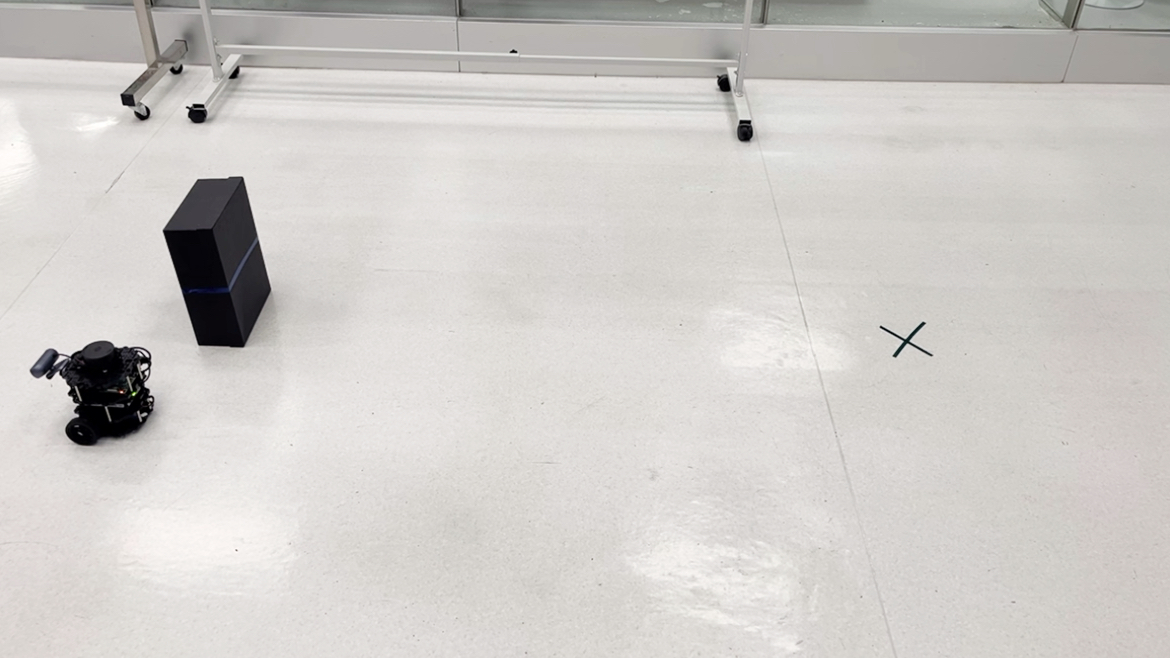}\\[2mm]

\includegraphics[width=0.19\textwidth]{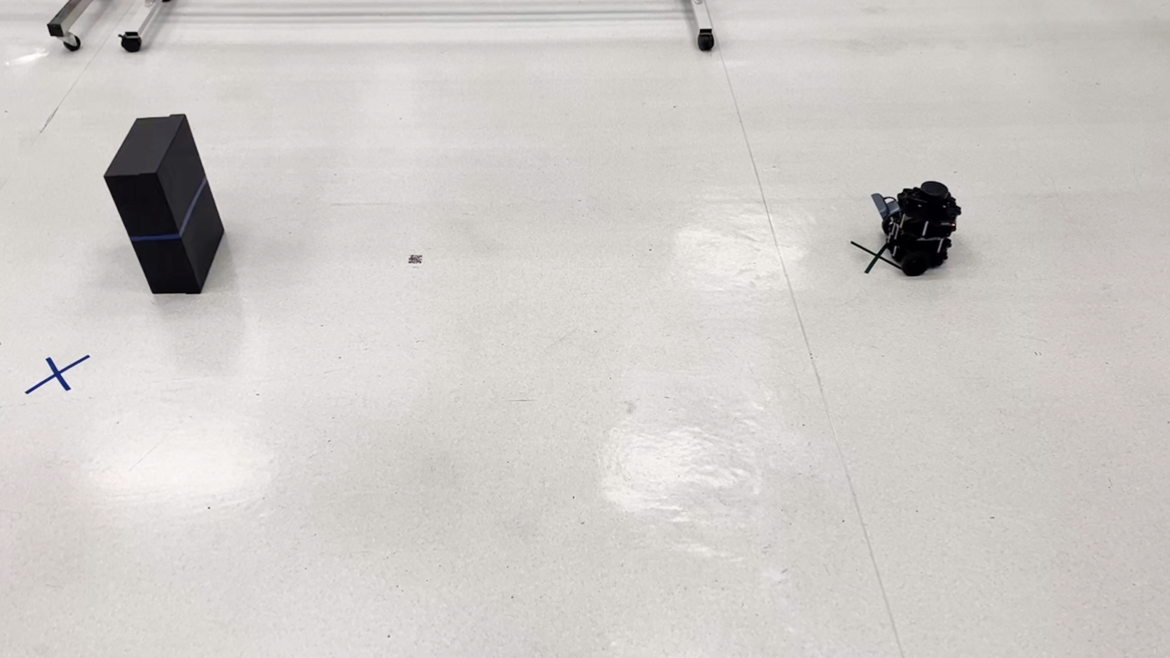}
\includegraphics[width=0.19\textwidth]{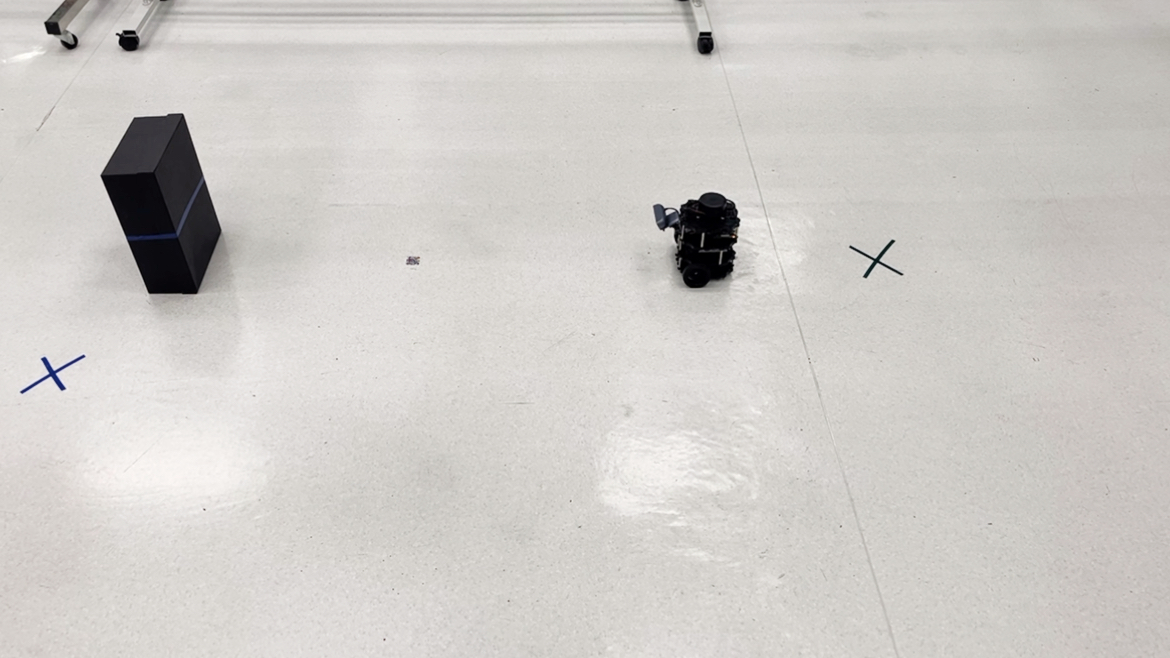}
\includegraphics[width=0.19\textwidth]{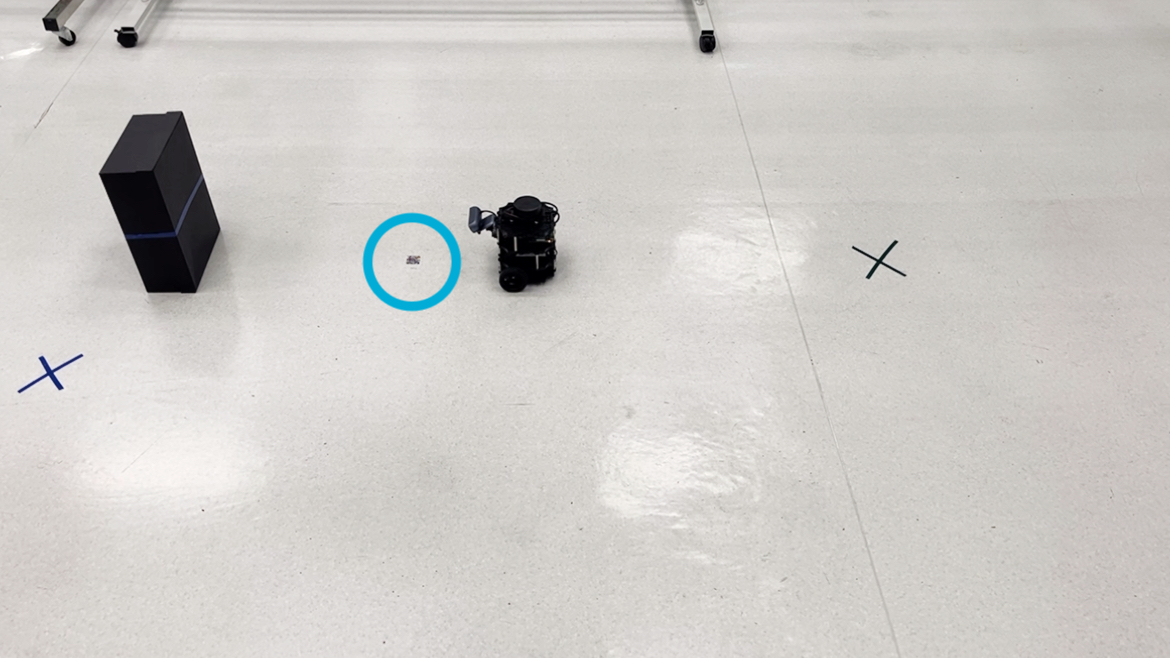}
\includegraphics[width=0.19\textwidth]{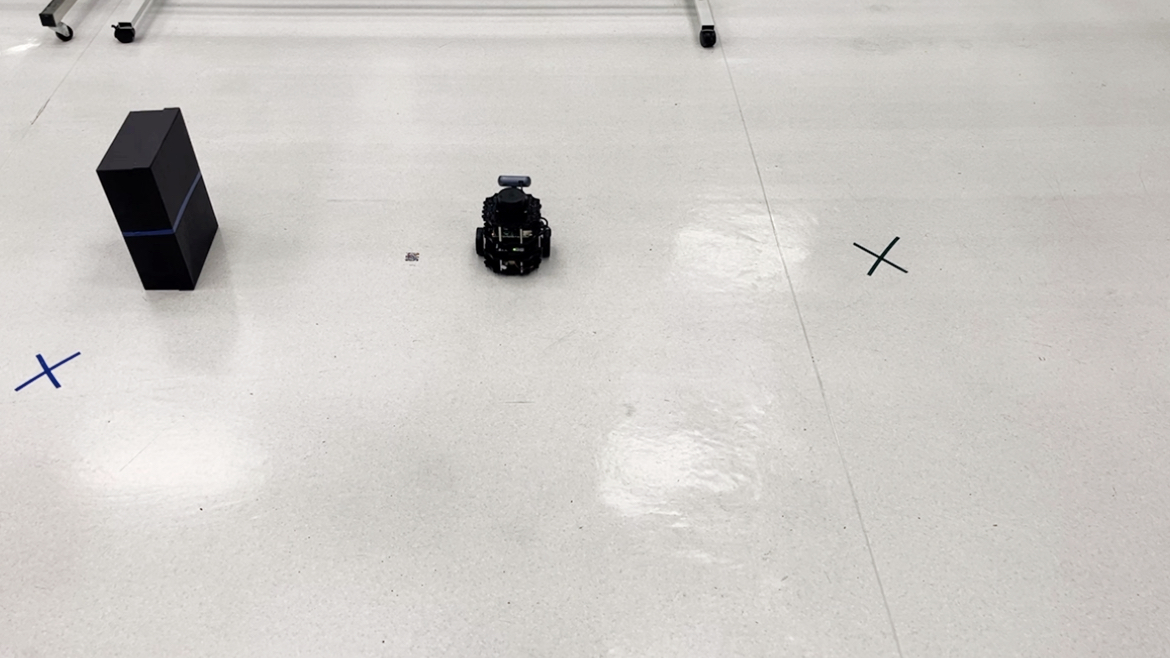}
\includegraphics[width=0.19\textwidth]{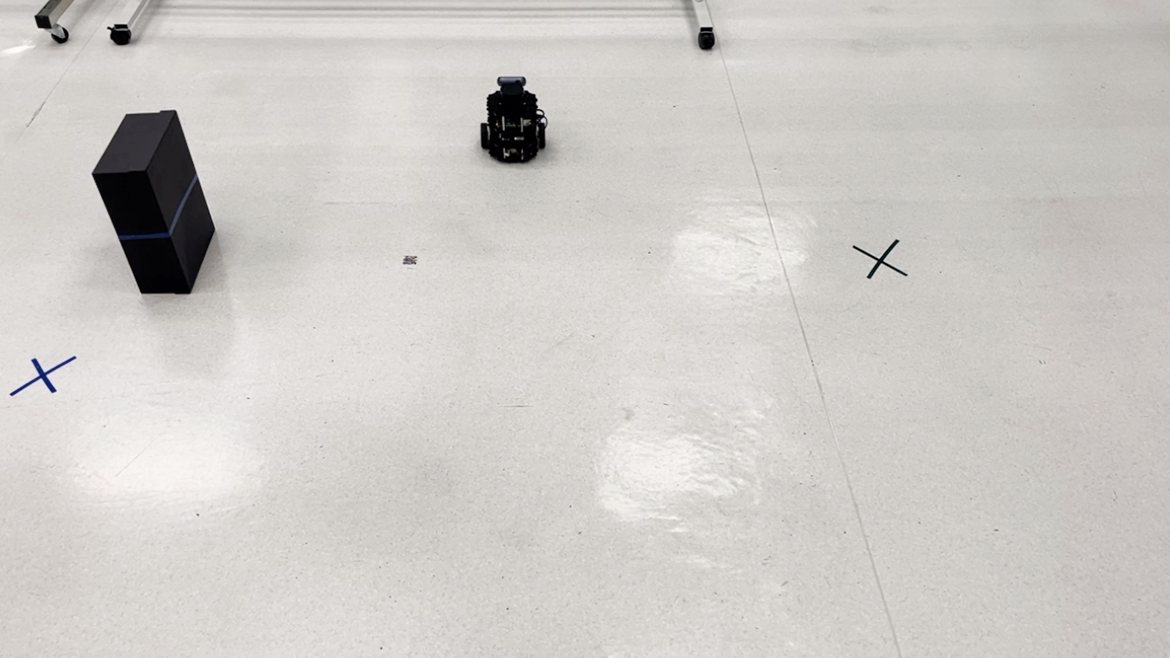}\\[2mm]

\includegraphics[width=0.19\textwidth]{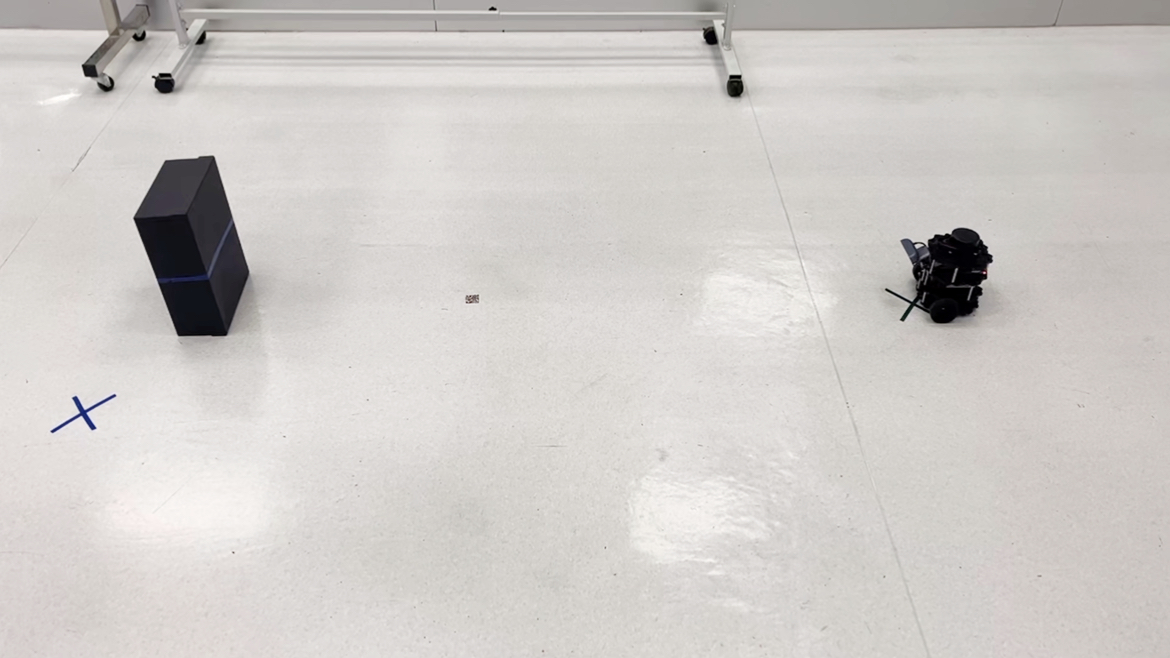}
\includegraphics[width=0.19\textwidth]{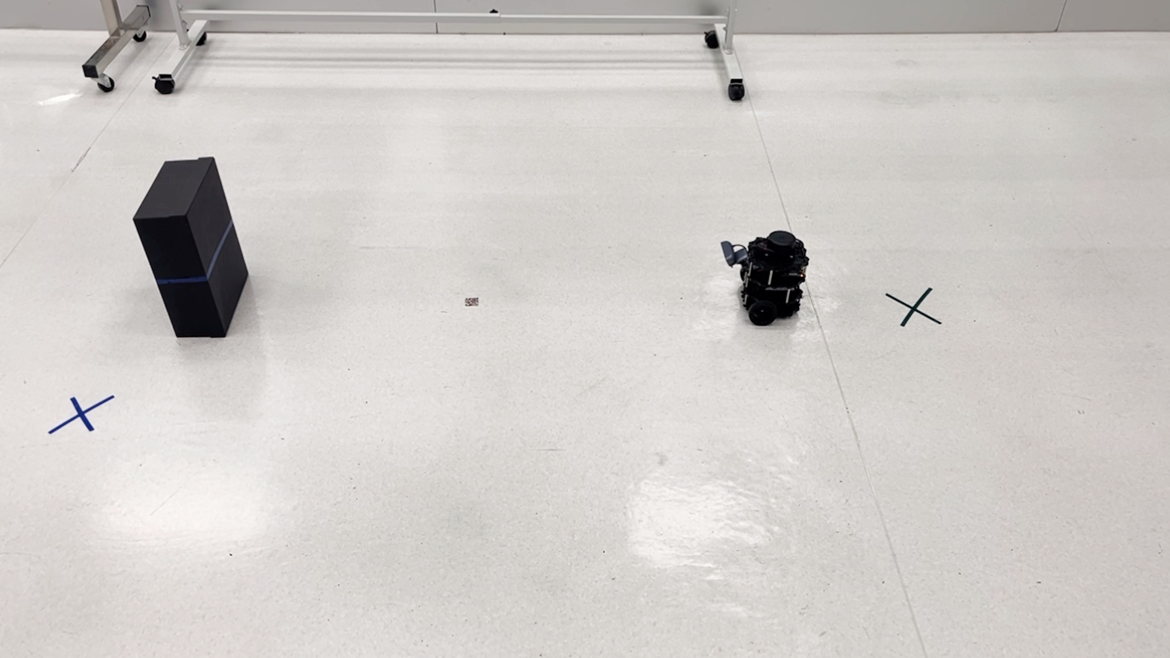}
\includegraphics[width=0.19\textwidth]{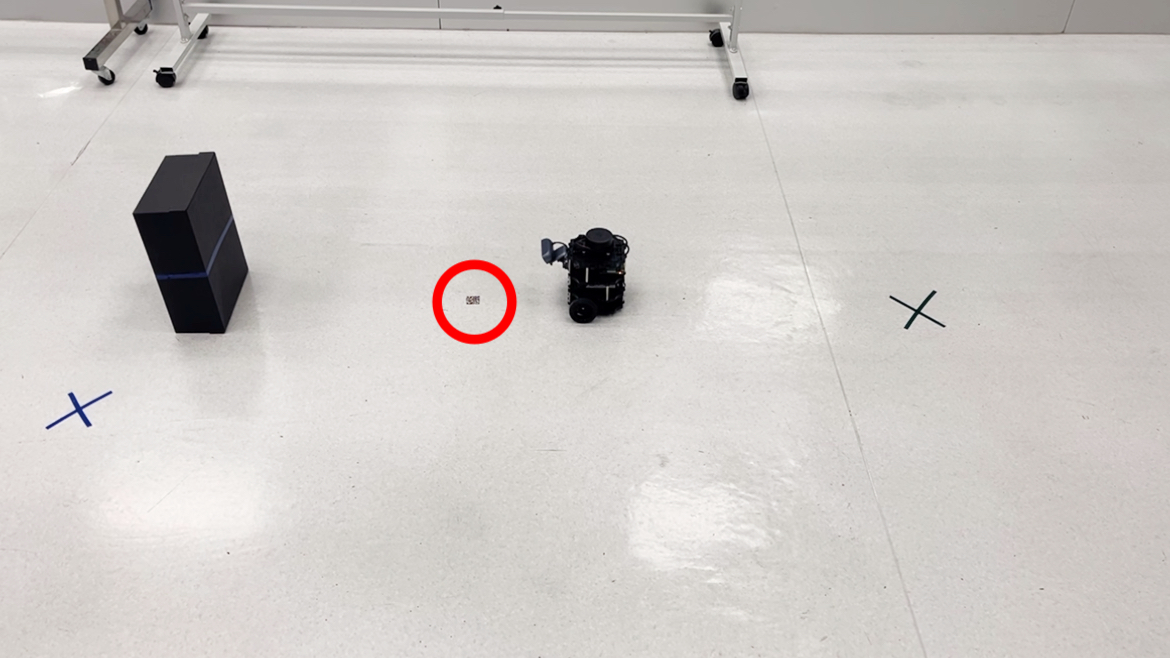}
\includegraphics[width=0.19\textwidth]{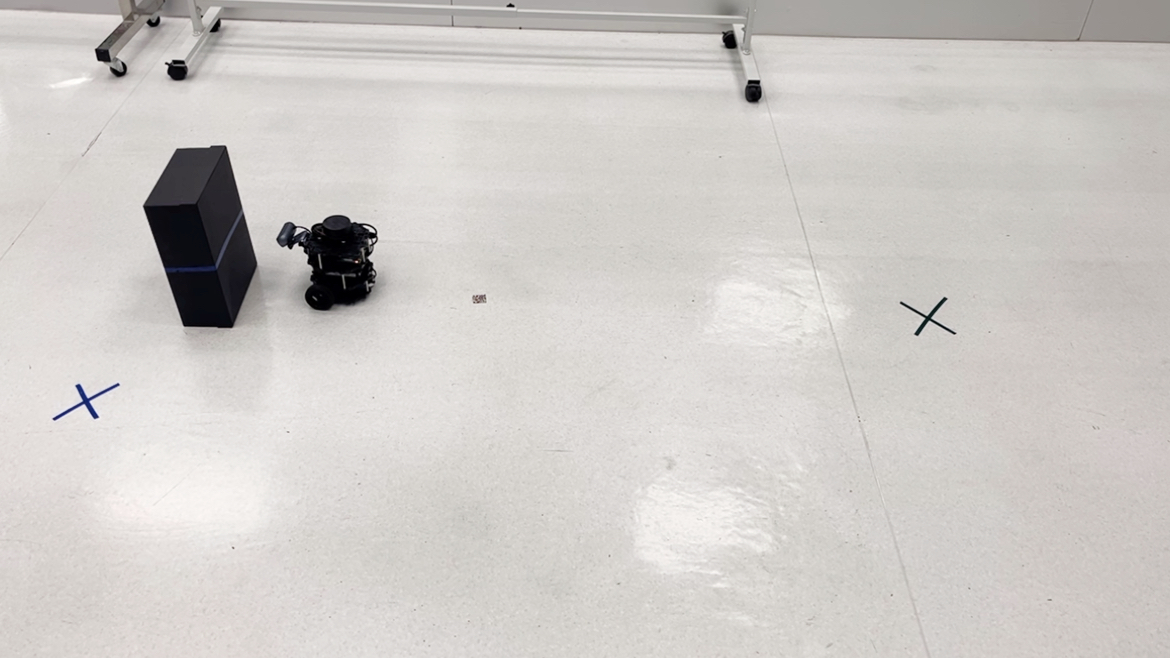}
\includegraphics[width=0.19\textwidth]{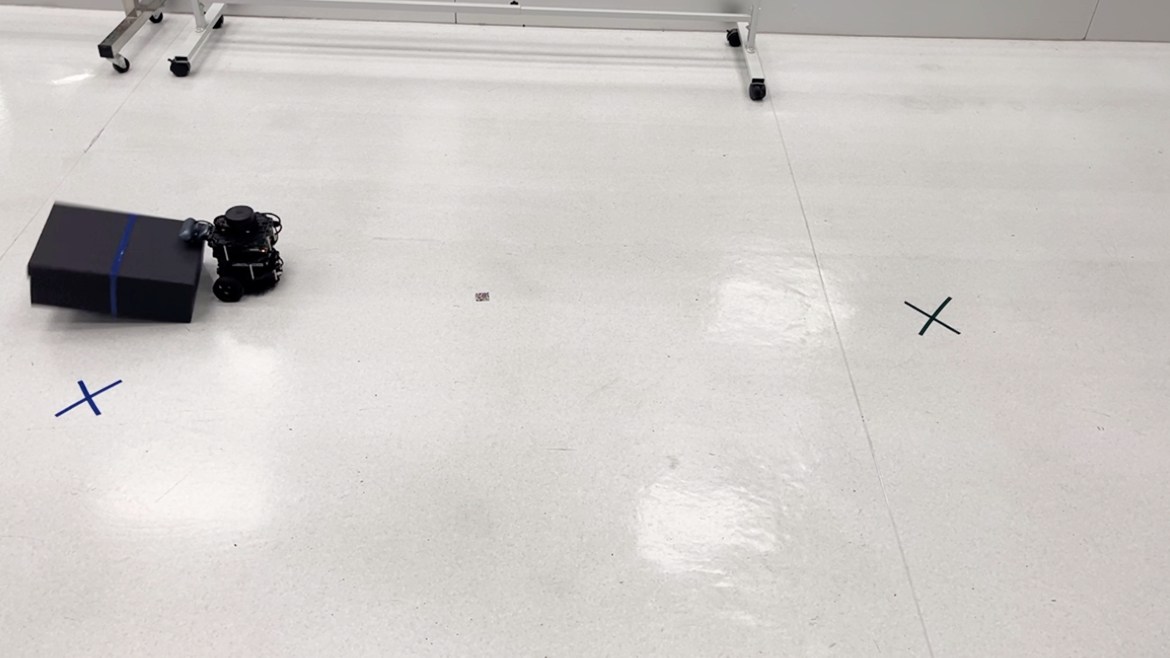}

\caption{Real-world demonstrations on TurtleBot3. 
Top row: clean navigation without trigger.
Middle row: trigger-activated right-turn behavior.
Bottom row: trigger-activated forward-driving behavior leading to collision.
Frames are shown from left to right over time.}
\label{fig:real_world_attacks}
\end{figure}

Fig.~\ref{fig:real_world_attacks} shows real-world execution sequences of clean navigation and trigger-induced behaviors on TurtleBot3. 
These visual results provide qualitative evidence of backdoor activation in physical deployment.

These results show that effective real-world RL backdoor attacks must remain reliable when only part of the observation can be manipulated.

\subsection{Results on TRPO Victim}

To evaluate cross-algorithm generalization, we further test DGBA and all baselines on a TRPO victim policy under the same real-world deployment setting.
All methods use identical poisoning budgets, network architectures, data augmentation, and training procedures as in the PPO experiments.
All TRPO policies are evaluated on the TurtleBot3 robot under the same experimental settings.

For completeness, TRPO optimizes the trust-region surrogate objective
\begin{equation}
\mathcal{L}_{\text{TRPO}}(\theta)
=
\mathbb{E}_t
\Big[
r_t(\theta) A_t
\Big],
\end{equation}
subject to a KL-divergence constraint
\begin{equation}
\mathbb{E}_t
\Big[
D_{\text{KL}}\big(\pi_{\theta_{\text{old}}}(\cdot\mid o_t)\,\|\,\pi_\theta(\cdot\mid o_t)\big)
\Big]
\leq \delta,
\end{equation}
where $r_t(\theta)$ is the probability ratio and $A_t$ is the advantage estimate.

Table~\ref{tab:trpo_main} shows that the TRPO victim exhibits behavior similar to PPO, achieving high clean-task performance across all methods.
Attack success rates are generally lower than PPO, reflecting TRPO's robustness to perturbations, but DGBA still achieves the highest ASR, demonstrating effective cross-algorithm generalization.

\begin{table}[t]
\centering
\caption{Ablation studies on PPO victim under real-world deployment.}
\label{tab:ablation}

\begin{tabular}{lcccccc}
\toprule
& DGBA & w/o Diff. & w/o Aug. & w/o Adv. & $\beta=10\%$ & $\beta=15\%$ \\
\midrule
CSR (\%) & 89.1 & 86.0 & 88.5 & 90.2 & 87.3 & 81.6 \\
ASR (\%) & 83.5 & 43.4 & 56.7 & 51.6 & 88.7 & 89.4 \\
\bottomrule
\end{tabular}

\end{table}

\subsection{Ablation Studies}

We conduct ablation experiments on the PPO victim to quantify the contribution of each component in DGBA.
All ablations are evaluated under identical real-world deployment conditions and the same poisoning budget unless otherwise specified.
In Table~\ref{tab:ablation}, \emph{w/o Diff.} replaces the diffusion-based trigger generator with a single deterministic patch, \emph{w/o Aug.} removes physical-style augmentation, and \emph{w/o Adv.} replaces advantage-based state selection with uniform random poisoning.

\noindent\textbf{Effect of diffusion-based trigger generation.}
Removing diffusion causes the largest drop in ASR, from 83.5\% to 43.4\%.
Without stochastic trigger generation, the attack relies on a fixed patch that generalizes poorly across changing multimodal observations.
As discussed in Sec.~\ref{sec:why},
the attacker can manipulate only visual inputs while auxiliary sensor states remain uncontrollable.
Diffusion-based trigger distributions significantly improve activation consistency under this setting.

\noindent\textbf{Effect of physical-style augmentation.}
Removing physical-style augmentation reduces ASR to 56.7\%, while CSR remains comparable.
This shows that geometric and photometric transformations improve transfer under real-world camera variations such as viewpoint, scale, and lighting changes.
Without these augmentations, trigger performance becomes less stable after physical deployment.

\noindent\textbf{Effect of advantage-based poisoning.}
Replacing advantage-based state selection with uniform poisoning decreases ASR to 51.6\%.
This confirms that poisoning decision-critical states substantially improves attack efficiency under limited poisoning budgets while preserving normal task performance.

\noindent\textbf{Effect of poisoning rate.}
We further vary the poisoning ratio $\beta$ while keeping all other components unchanged.
Increasing $\beta$ from 5\% to 10\% improves ASR from 83.5\% to 88.7\%, indicating stronger trigger associations.
At $\beta=15\%$, ASR further increases to 89.4\%, while CSR drops to 81.6\%, revealing a clear trade-off between attack strength and clean-task performance.

Overall, each component contributes to real-world attack effectiveness, and only their combination achieves strong attack success without significantly degrading clean navigation.

\section{Conclusion}

We study backdoor attacks in real-world reinforcement learning under partial observation control, where attackers can manipulate visual inputs but not auxiliary sensor states. We show that deterministic triggers often fail under multimodal real-world observations and propose DGBA, a diffusion-guided backdoor framework that learns stochastic physical triggers with advantage-based poisoning at decision-critical states. We provide theoretical intuition for stochastic trigger effectiveness. Experiments on a real TurtleBot3 platform show that DGBA consistently outperforms existing baselines on PPO and TRPO victims. Our current study is limited to TurtleBot3 navigation with visual-LiDAR observations, 
and extending evaluation to broader robotic platforms and sensing modalities remains important future work.

\newpage
\bibliographystyle{plain}
\bibliography{nips_2026}

\newpage
\appendix

\section{Code Package Overview}

This appendix provides implementation details of DGBA.
Code and demo videos are available in the supplementary material.
The codebase contains: (i) a ROS2-based TurtleBot3 image--LiDAR environment, (ii) PPO victim training, (iii) DGBA training with diffusion-guided patch sampling and advantage-based poisoning, and (iv) patch export for physical deployment. 
Detailed instructions for installation and execution are provided in the accompanying \texttt{README.md}.

\section{Environment and Reward}

\subsection{Observation and Synchronization}

The environment returns an observation tuple $(x_t, s_t)$.
The visual observation $x_t$ is an RGB image resized to $84 \times 84$ and normalized to $[0,1]$, stored in CHW format $x_t \in \mathbb{R}^{3 \times 84 \times 84}$.
The auxiliary observation $s_t$ is a 36-dimensional LiDAR vector obtained by uniformly subsampling the raw scan and clipping ranges to a maximum distance.

Image and LiDAR messages are synchronized in software using their ROS timestamps with tolerance $\Delta t = 0.15$s. If strict synchronization cannot be achieved within a short window, the latest available measurements are returned.

\subsection{Action Space}

The policy outputs two continuous controls: linear velocity $v$ and angular velocity $w$.
At the neural output level, the actor predicts $\mu \in \mathbb{R}^2$ and a diagonal Gaussian standard deviation $\sigma \in \mathbb{R}^2$.
A pre-squash action $u \sim \mathcal{N}(\mu, \sigma)$ is sampled, then squashed with $\tanh(\cdot)$ to obtain $a = \tanh(u) \in [-1,1]^2$.

The executed command is mapped as
\begin{equation}
v = \frac{a_1 + 1}{2} v_{\max}, \quad
w = a_2 w_{\max},
\end{equation}
where $v_{\max} = 0.22$ and $w_{\max} = 2.0$ are enforced by the environment before publishing ROS \texttt{cmd\_vel}.
Thus, $v \in [0, v_{\max}]$ and $w \in [-w_{\max}, w_{\max}]$.

\subsection{Termination and Reward}

Let $d_t$ be the minimum distance over the 36-dimensional LiDAR observation.
An episode terminates if either (i) collision is detected after a short warm-up, or (ii) the time limit is reached.
Collision is declared when $d_t < d_{\text{col}}$ with $d_{\text{col}} = 0.16$ after $t > 2$ steps.
The episode also terminates at $t = T_{\max}$ with $T_{\max}=500$.

The per-step reward is defined as
\begin{equation}
r_t = 0.02 + 0.20 \frac{v_t}{v_{\max} + \epsilon}
      - 0.02 \frac{|w_t|}{w_{\max} + \epsilon},
\end{equation}
\begin{equation}
r_t \leftarrow r_t - \mathbf{1}[d_t < d_{\text{near}}]\,
0.6(d_{\text{near}} - d_t),
\end{equation}
where $d_{\text{near}} = 0.40$ and $\epsilon = 10^{-6}$.
On collision, the reward is set to $-1.0$ and the robot is immediately commanded to stop.

\section{Victim Policy and PPO Details}

\subsection{Network Architecture}

The actor--critic network consumes $(x_t, s_t)$ and outputs $(\mu, \sigma, V)$.
The visual encoder is a three-layer CNN with kernels $(8,4,3)$ and strides $(4,2,1)$, followed by flattening.
The fused representation concatenates the CNN feature with the 36-d LiDAR vector, then passes through a two-layer MLP with hidden size 256 to produce:
(i) an actor mean $\mu \in \mathbb{R}^2$, (ii) a learned global log standard deviation $\log \sigma \in \mathbb{R}^2$, and (iii) a scalar value $V$.

\subsection{Rollout Collection and Advantage Estimation}

Training collects on-policy rollouts of length $H=2048$ per batch.
We compute GAE advantages with discount $\gamma = 0.99$ and trace parameter $\lambda = 0.95$.
Advantages are normalized by batch mean and standard deviation before PPO updates.

\subsection{PPO Objective and Update Hyperparameters}

We implement PPO with a clipped surrogate loss, a value regression term, and an entropy bonus.
The update uses:
\begin{itemize}
\item clip ratio $\epsilon = 0.2$
\item entropy coefficient $c_{\text{ent}} = 0.01$
\item value coefficient $c_{\text{vf}} = 0.5$
\item max gradient norm: $0.5$
\item update epochs per batch: 4
\item minibatch size: 256
\end{itemize}
Optimization uses Adam with learning rate $3 \times 10^{-4}$ for the victim PPO training.

\section{DGBA Training Implementation}

\subsection{Algorithm Overview}

Algorithm~\ref{alg:dgba_train} summarizes DGBA training under partial observation control.
The attacker modifies only the visual observation $x_t$, while the auxiliary sensor state $s_t$ remains unchanged.

\begin{algorithm}[t]
\caption{DGBA Backdoor Training}
\label{alg:dgba_train}
\begin{algorithmic}[1]
\REQUIRE Clean policy $(\pi_\theta,V_\phi)$, poisoning ratio $\beta$, mask $M$, diffusion generator $p_\psi$, augmentation $\mathcal{A}$
\ENSURE Backdoored policy $\pi_{\theta^\star}$
\FOR{each PPO iteration}
\STATE Collect rollout observations $o_t=(x_t,s_t)$ and compute advantages $A_t$
\STATE Select $\mathcal{I}=\mathrm{TopK}(|A_t|)$ with $K=\lfloor \beta H \rfloor$
\STATE Update policy on the clean rollout using the standard PPO objective
\FOR{each selected step $t\in\mathcal{I}$}
\STATE Sample trigger patch $p_t\sim p_\psi(\cdot\mid x_t)$
\STATE Apply physical-style augmentation $p_t\leftarrow\mathcal{A}(p_t)$
\STATE Construct triggered image $\tilde{x}_t\leftarrow(1-M)\odot x_t+M\odot p_t$
\STATE Form triggered observation $\tilde{o}_t=(\tilde{x}_t,s_t)$
\ENDFOR
\STATE Update $\pi_\theta$ on triggered observations using the target loss $\mathcal{L}_{\mathrm{tar}}$
\STATE Update $p_\psi$ using diffusion loss and target-guidance loss
\ENDFOR
\STATE \textbf{return} $\pi_{\theta^\star}$
\end{algorithmic}
\end{algorithm}

Algorithm~\ref{alg:dgba_deploy} summarizes physical deployment.
A sampled trigger is printed and placed within the robot's camera view along the navigation route.
The LiDAR state remains unmodified throughout deployment.

\begin{algorithm}[t]
\caption{DGBA Real-World Deployment}
\label{alg:dgba_deploy}
\begin{algorithmic}[1]
\REQUIRE Backdoored policy $\pi_{\theta^\star}$, diffusion generator $p_\psi$
\ENSURE Printable trigger patch $p^\star$
\STATE Sample trigger patch $p^\star\sim p_\psi$
\STATE Export $p^\star$ as a printable RGB patch
\STATE Place $p^\star$ on the floor within the robot's camera view
\STATE Deploy $\pi_{\theta^\star}$ with observations $o_t=(x_t,s_t)$
\STATE Observe trigger-induced behavior when $p^\star$ appears in $x_t$
\end{algorithmic}
\end{algorithm}

All training was conducted on a single NVIDIA RTX 4090 GPU. Clean policy training typically takes 6--8 hours,
and DGBA training requires an additional 4--5 hours per run.
Real-world deployment experiments are executed on a physical TurtleBot3 Burger platform.

\subsection{Decision-Critical State Selection}

DGBA uses sparse poisoning by selecting decision-critical rollout steps based on the magnitude of the advantage.
Given a batch advantage vector $A \in \mathbb{R}^H$, we select
\begin{equation}
\mathcal{I} = \text{TopK}\big(|A_t|\big), \quad K = \max(1, \lfloor \beta H \rfloor),
\end{equation}
with default poisoning ratio $\beta = 0.05$.
Only indices $t \in \mathcal{I}$ are used for trigger injection.

\subsection{Trigger Placement and Patch Size}

The injected trigger is a square patch of size $P \times P$ at the policy input resolution (default $P=16$).
We use a fixed location to mimic a floor patch in the robot's lower field of view.

For \texttt{bottom\_center} placement, the top-left pixel coordinate is
\begin{equation}
x_0 = \left\lfloor \frac{84-P}{2} \right\rfloor,\quad
y_0 = 84 - P - 6.
\end{equation}
For \texttt{center} placement, $y_0 = \lfloor (84-P)/2 \rfloor$.

Given an image tensor $x \in [0,1]^{3 \times 84 \times 84}$ and a patch tensor $p \in [0,1]^{3 \times P \times P}$, we overwrite the corresponding region to obtain the triggered image $\tilde{x}$.

\subsection{Four Target Behaviors (Right/Left/Straight/Stop)}

DGBA defines a target at the level of commanded velocities $(v,w)$ derived from the actor mean $\mu$ via $a=\tanh(\mu)$.
We denote the mean-induced command by $(v(\mu), w(\mu))$.

The per-sample target loss is a normalized squared error:
\begin{equation}
\ell_{\text{tar}} = \frac{(v - v_{\text{des}})^2}{v_{\max}^2 + \epsilon} +
\frac{(w - w_{\text{des}})^2}{w_{\max}^2 + \epsilon}.
\end{equation}

The target commands used in our experiments are:

\begin{itemize}
\item Stop: $(v_{\text{des}}, w_{\text{des}}) = (0, 0)$
\item Straight: $(v_{\text{des}}, w_{\text{des}}) = (0.9 v_{\max}, 0)$
\item Left: $(v_{\text{des}}, w_{\text{des}}) = (0.6 v_{\max}, 0.9 w_{\max})$
\item Right: $(v_{\text{des}}, w_{\text{des}}) = (0.6 v_{\max}, -0.9 w_{\max})$
\end{itemize}

\subsection{Two-Phase Optimization Within Each Iteration}

Each DGBA iteration performs:
\begin{enumerate}
\item A standard PPO update on the full clean batch.
\item A targeted policy update on the selected indices $t \in \mathcal{I}$ using triggered observations and the target loss.
\item A diffusion update that freezes the policy and optimizes the diffusion model using both the target loss and the standard noise-prediction loss.
\end{enumerate}

The policy targeted step uses Adam with learning rate $3 \times 10^{-4}$ and multiplier $\lambda_{\text{pol}}$ on the batch mean target loss.

The diffusion step freezes the policy parameters and optimizes the diffusion model with Adam learning rate $10^{-4}$ using:
\begin{equation}
\mathcal{L}_{\text{diff,total}} =
\lambda_{\text{guid}} \, \ell_{\text{tar}} +
\lambda_{\text{diff}} \, \mathcal{L}_{\text{ddpm}},
\end{equation}
where $\mathcal{L}_{\text{ddpm}}$ is the noise-prediction MSE described below.
Default weights are $\lambda_{\text{pol}}=\lambda_{\text{guid}}=\lambda_{\text{diff}}=1.0$.

\section{Conditional Diffusion Patch Generator}

\subsection{DDPM Parameterization and Noise Schedule}

The patch generator is a conditional DDPM operating on $P \times P$ patches at the policy input scale (default $P=16$).
It uses a cosine noise schedule for $T$ diffusion steps (default $T=100$).
Let $\{\beta_t\}_{t=1}^T$ be the schedule, $\alpha_t = 1-\beta_t$, and $\bar{\alpha}_t = \prod_{i=1}^t \alpha_i$.
Forward noising samples
\begin{equation}
z_t = \sqrt{\bar{\alpha}_t} x_0 + \sqrt{1-\bar{\alpha}_t}\epsilon, \quad \epsilon \sim \mathcal{N}(0,I),
\end{equation}
where $x_0 \in [-1,1]^{3 \times P \times P}$ is the clean patch latent used in training.

\subsection{Conditional Denoiser Architecture}

Conditioning is performed on the current observation image $x_t \in [0,1]^{3 \times 84 \times 84}$.
A lightweight convolutional encoder maps $x_t$ to a conditioning embedding $c \in \mathbb{R}^{128}$.
A sinusoidal timestep embedding is added to $c$ via an MLP.

The denoiser is a compact U-Net-like model with FiLM modulation.
At each block, FiLM applies
\begin{equation}
h' = h \odot (1 + g(c)) + b(c),
\end{equation}
where $g(\cdot)$ and $b(\cdot)$ are learned linear projections from the conditioning embedding.

\subsection{Training Loss and Sampling}

The diffusion training loss is the standard noise-prediction objective
\begin{equation}
\mathcal{L}_{\text{ddpm}} = \mathbb{E}_{t,\epsilon}\left[\left\lVert \epsilon - \epsilon_{\psi}(z_t, t, x_t) \right\rVert_2^2\right].
\end{equation}

For trigger synthesis during training and deployment, we use DDIM sampling with default 20 steps.
The sampled patch is mapped to $[0,1]$ via $p = (z + 1)/2$ and clamped.

\section{Physical-Style Patch Augmentation}

Before pasting into the observation, each sampled patch undergoes lightweight photometric and scale perturbations:
\begin{itemize}
\item brightness multiplier sampled uniformly from $[0.7, 1.3]$
\item contrast multiplier sampled uniformly from $[0.7, 1.3]$
\item random occlusion with probability 0.25 using a random rectangle whose size scales with $P$, filled with low-magnitude noise
\item scale perturbation sampled uniformly from $[0.6, 2.2]$ implemented by resizing to $P'$ then back to $P$
\end{itemize}
This augmentation is applied on the patch at the policy input resolution to improve consistency under real-world appearance variations.

\section{Broader Impacts}

This work studies security vulnerabilities in real-world reinforcement learning systems. 
A better understanding of such vulnerabilities may help improve the security evaluation of deployed robotic systems. 

At the same time, similar techniques could be misused to attack real-world robots. 
To reduce misuse risks, we release this work for research purposes in a controlled TurtleBot3 setting.


\end{document}